# Preprint: ARPPS：Augmented Reality Pipeline Prospect System

Xiaolei Zhang[1], Yong Han[1,*], DongSheng Hao[1], Zhihan Lv[2]

[1] Engineering Research Center of Marine Information Technology, College of Information Science and Engineering, Ocean University of China, Qingdao, P. R. China

[2] Shenzhen Institutes of Advanced Technology (SIAT), Shenzhen, P. R. China

**Abstract.** This is the preprint version of our paper on ICONIP. Outdoor augmented reality geographic information system (ARGIS) is the hot application of augmented reality over recent years. This paper concludes the key solutions of ARGIS, designs the mobile augmented reality pipeline prospect system (ARPPS), and respectively realizes the machine vision based pipeline prospect system (MVBPPS) and the sensor based pipeline prospect system (SBPPS). With the MVBPPS's realization, this paper studies the neural network based 3D features matching method.

## 1 Introduction

Since the human activities are related with the geographic information all the time, it is very natural and extremely important to integrate the augmented reality technology with the GIS. Since its generation, the augmented reality has gone through several development phases: computer indoor augmented reality, helmet-type outdoor augmented reality and mobile based outdoor augmented reality. Its application is gradually converted from the traditional exhibition to the application combined with the industry. This paper studies the mobile based outdoor augmented reality geographic information system (ARGIS) [1-3] and analyzes it by combining the examples for underground pipeline prospect. In the process of the realization of the examples, this paper also studies the neural network based 3D features matching method to optimize the feature matching process.

The traditional GIS application is displayed on the map in the form of 2D symbols after analyzing the result based on the 2D maps. However, due to the lack of strong intuition, it is difficult for the user with weak geographic knowledge to distinguish it; even the field experts also have some difficulties in distinguishing in a complicated environment. For recent years, with the 3DGIS and VRGIS development, the 3D virtual model of real environment was established to replace the 2D map [4], which improves the intuition of marker distinguishing and promotes the visualization development of GIS. However, the 3D virtual model has high requirements on hardware while it still has difference with the real world scene. As a matter of fact, the real world scene is a complicated and perfect "3D space map" for itself. If the real world scene is taken as the "map" for GIS application operation and the inquired and analyzed result is displayed on the real world scene via virtual information to achieve the interaction between the virtual space geographic information and the real world scene, so the GIS application experience can be enhanced.

Carry out scene augmentation [5,6] for human's visual system by organically integrating the 2D or 3D pictures, text notes and other virtual information generated by the computer into the real world scene which can be seen by the user. The augmented reality has the effect of expressing the sense of reality but cannot store and control data. However, GIS has the functions of storage, management and analysis of space information, which precisely makes up for this deficiency. The integration of them can provide the user with the location-based service and can combine all kinds of virtual space information stored in the GIS space data with the scene actually observed [7]. Besides, such integration can not only enhance the GIS user's sense of real environment and the interactive experience but also supplies a new means for the research on GIS visualization.

Sun M[8] (2004) brought forward the concept of ARGIS(Augmented reality geographic information system), a geographic information system, digitally describes, stores and controls the objective geographic world, meanwhile, integrates such descriptions into the real world, offers the space information of a designated object and supplies the outdoor mobile information interaction. Guo Y and others (2008) pointed out that the significance of ARGIS is to apply the mobile computing and augmented reality technologies to the traditional space information service to change the traditional location-based service mechanism; then, human, which is the subject, the real world, which is the object, and the digital world transmitted by internet can combined with each other seamlessly so as to realize the interaction without being limited by any time and space and to alter the human-to-digital world and human-to-real world interactive pattern, which will provide the space-based industry system with a brand-new industrial pattern[9].

Gethin D's article [10] of 2002 adapted GPS/INS for 3D registration of outdoor geographic information, realizing the visual presentation of outdoor underground facility structure. The subsurface data visualization system of University of Nottingham, which uses the GPS/ INS integrated system in system registration, can carry out the 3D visualization for all kinds of the following subsurface characteristics such as geologic structure, underground pipe system, underground land pollution zone etc. Gerhard Schall has published 15 articles on outdoor augmented reality since 2008, emphasizing the application of hardware sensor. One of these articles written in 2009 [11] designs an underground facility visualization augmented reality system frame and an integrated-form handheld device and integrates GPS, camera, wireless network and other devices but is not the mobile-phone-based application. The 3D registration uses the pure hardware sensor technology (GPS+ inertia measurement equipment). Gerhard Schall started to research the visual and sensor hybrid tracking registration method (12) in 2010 and obtained a series of research achievements [13, 14, 15]. The article [16] designs an industrial solution, using GPS positioning and CAD pipeline data in the real environment; it highly depends on the accuracy of GPS and there is no explanations about the orientation, tracking and other technical details [17-36].

## 2    Data description

Based on ARGIS framework, this paper achieves the mobile terminal augmented reality application system for urban underground pipeline prospecting (ARPPS). The system takes the underground pipeline data in Qingdao (500 square kilometers of construction area and 10 million population) as the experimental data to have achieved the machine vision based underground pipeline prospect system (MVBPPS) and sensor based underground pipeline prospect system (SBPPS).

Shape-format vector pipeline data is totally 2.7G. According to geometrical characteristics, data are divided into two types: pipeline and pipe point; according to the functional attributes, data are divided into 13 types including covered channel, power line carrier, power supply pipeline, monitoring signal pipeline, street lamp pipeline, hot water pipeline, drinking water (feed water) pipeline, natural gas pipeline, communication pipeline, sewage pipeline, rainwater pipeline, integrated pipeline, reclaimed water pipeline. Data are described as follows:

PipePoint: {OBJECTID, Pipe Point number, x coordinate x, y coordinate, ground elevation, characteristics, attached objects, burial depth of well bottom, type of well lid, specification of well lid, materials of well lid, offset distance, rotation angle}.

PipeLine: {OBJECTID id of starting point, id of ending point, burial depth of starting point, burial depth of ending point, elevation of starting point, elevation of ending point, x coordinate of starting point, y coordinate of starting point, x coordinate of ending point, y coordinate of ending point, material, way of burying, line type, pipe diameter, pipe length}.

## 3    Methods

### 3.1 Spatial data organization and distribution

As for the Shape-format vector pipeline data, the spatial database is imported using ArcGIS; the PostGIS open source database is adopted as the spatial database; the spatial data network services interface satisfying OGC standards is researched and developed independently. The data services interface receives the parameters within the visional field of the camera at client side, and the visional field of camera is expressed with a rectangle, and the range of visional field is the coordinates of the lower left corner and upper right corner of the rectangle [Lon_Min, Lat_Min, Lon_Max, Lat_max]; parameters are transmitted in the form of URL parameters under http protocol (range of visional field obtained by calculation at client side). These parameters are used for pipeline data retrieval, and results are returned to client side in the GeoJSON format.

### 3.2 3D registration and tracking

**MVBPPS:** The SFM (Structure from Motion) technology is adopted to achieve the 3D registration and tracking. SFM is a kind of registration and tracking method based on natural scene reconstruction and is a kind of on-line reconstruction method. Systems mentioned in this paper utilize the PTAM (Parallel Tracking and Mapping) system, can process thousands of natural feature points in a real-time way under the PC environment, and have good performance in the aspects of precision and robustness. Klein [40] and other persons proposed the on-line reconstruction method to solve the problem of natural scene registration in mobile augmented reality, and they revised the PTAM system to achieve its operation in mobile equipment; experimental results indicate that the revised system can achieve the reconstruction and registration of small scenes, although its precision and robustness reduce slightly.

PTAMM(Parallel Tracking and Multiple Mapping)[49] extends PTAM system to allow it to use multiple independent cameras and multiple maps. This allows maps of multiple workspaces to be made and individual augmented reality applications associated with each. As the user explores the world the system is able to automatically relocate into previously mapped areas.

Recently, many scholars have used neural network to optimize the computing process of features matching. Transiently chaotic neural network (TCNN) [50] is one of the most representative. TCNN exploiting the rich behaviors of nonlinear dynamics have been developed as a new approach to extend the problem solving ability of standard HNN [51]. The TCNN model can be presented as follows:

$$x_i(t) = \frac{1}{1 + e^{-y_i(t)(1+\varepsilon)}} \tag{1}$$

$$y_i(t+1) = ky_i(t) + \alpha\left(\sum_{j=i,j\neq ij}^{n} w_{ji}x_j(t) + I_i\right) - z_i(t)(x_i(t) - I_0) \tag{2}$$

$$z_i(t+1) = (1-\beta)z_i(t) \tag{3}$$

where (i=1,2,..., n) , $x_i$ = output of neuron i, $y_i$ = internal state of neuron i, $w_{ij}$ = connection weight from neuron j to neuron i, $w_{ij} = w_{ji}$ , $I_i$ = input bias of neuron i, α = positive scaling parameter for inputs, k = damping factor of nerve membrane (0 ≤ k ≤ 1) , $z_i(t)$ = self-feedback connection weight (refractory strength) ≥ 0, β= damping factor of $z_i(t)$, 0 < β < 1, $I_0$ = positive parameter, ε = steepness parameter of the output function (ε > 0) .

When matching the image features, the neural network structure is a 2D array. If the reference image is represented by a set of feature points G, which size is M; Then an input scene image that consist of several overlapping objects, can be represented as another set of feature points S, which size is N. The number of neurons in the network will be MN. The output status of the neural $v_{ij}$ represents the matching state of point No. i, and No. j. If i matched j, $v_{ij}$ will be set to 1, else $v_{ij}$ will be set to 0. [52]

The common method for SFM is feature based approach (see [41] for more details). The feature base SFM depends on robust feature detection and matching, geometric image transformation, image stitching and adjustment. For robust feature detection and matching in two corresponding images, SIFT feature [42] detector with FLANN [43] match is normally employed to extract the key points in a given video sequence. It has known [44] that the hymnographies induced by the plane $n^T X + d = 0$ under the coordinates $\Pi_E = (n^T, d)^T$ is:

$$H_{ij} = K_i\left[R - \frac{tn^T}{d}\right]K_j^{-1} \tag{4}$$

And if the camera rotates about its optical center, the group of transformations the images may undergo is a special group of hymnographies [45], [46]:

$$H_{ij} = K_i R_i R_j^T K_j^{-1} \tag{5}$$

**SBPPS:** Gyroscope, GPS, electronic compass, accelerometer and other sensor data are integrally used and fused to achieve 3D registration and tracking of virtual objects.

The difference between the 3D registration process and the traditional augmented reality registration process lies in determination of spatial range of visualized pipeline data. In the process of determining the spatial range of visualized pipeline data, the positioning parameters [x, y] of GPS and the loading radius r are utilized; spatial range parameters of rectangle are [x-r,y-r,x+r,y+r], in which x is longitude, y is latitude and r is the distance value of 10m range in the World Coordinate System. After determination of spatial range for pipeline, data are obtained through service invocation at the server side, and are rendered to the screen or projector through coordinate transformation.

The process of tracking is achieved by adopting the angular acceleration of gyroscope and accelerometer as well as the direction parameters of compass to perform the real-time monitoring and to inversely calculate the position alteration of the virtual object. Currently, the problems of this method are the shaking and drifting phenomena due to the limited precision of mobile sensor. In this experiment, the weighed recursive average filtering algorithm [47] is adopted for data processing and shaking prevention. The weighed recursive average filtering algorithm is a kind of low-pass filtering algorithm; take the accelerometer as an example, and the corresponding parameters of the three coordinates x, y and z are Ax, Ay and Az respectively. Firstly, define the distance between new and old parameters as the standard for the degree of parameter shaking; new parameters are defined as $A'_x$, $A'_y$, $A'_z$.

$$d = \sqrt{(A'_x - A_x)^2 + (A'_y - A_y)^2 + (A'_z - A_z)^2} \qquad (6)$$

Then, take the distance as the independent variable to obtain the weighting coefficient α of low-pass filter; the value of α is obtained based on the range segmentation of d, as shown in the formula below:

$$\alpha(d) = \begin{cases} 0.001, d < low \\ 0.6, low \leq d < high \\ 0.9, d \geq high \end{cases} \qquad (7)$$

Finally, multiply the parameter variation value with the factor weighting coefficient in the weighed form and superpose it to the before-change data to obtain the new parameter value [48], as shown below:

$$\begin{aligned} A_x &= A_x + \alpha(d)(A'_x - A_x) \\ A_y &= A_y + \alpha(d)(A'_y - A_y) \\ A_z &= A_z + \alpha(d)(A'_z - A_z) \end{aligned} \qquad (8)$$

After the filtering processing of sensor signal, the shaking is reduced to some extent, and users' experience effect is improved; however, due to the limited precision of mobile sensor, simply adopting a low-pass filtering method cannot thoroughly solve the problem.

3.3 Mapping the sense of reality on virtual-real fusion

In this paper, the virtual sectioning is adopted for both MVBPPS and SBPPS, to achieve the mapping of the sense of reality on fusion between underground pipeline and ground; MVBPPS adopts the all-sight rectangular tunnel sectioning, while SBPPS adopts the 180 °front-sight circular tunnel sectioning. As for achieving the real fusion between grounds and sectioning tunnel, the OpenGL ES 2.0 mixed technology is utilized to achieve the transparent sheltering and to make the tunnel have a good immersion feeling on ground.

# 4    Operational results

Operating environment of application is nexus 5 mobile, and operating system is android 4.4/5.0.1. Data server: i5-3470 CPU, 16G memory, window server 2013.

**MVBPPS** is subjected to multi-environment tests, and operational results are tested in different lighting environments and under different ground texture conditions; testing environments include: sufficient sunshine at noon, twilight in the afternoon, floor tile covered parking lot, bituminous street, lawn, and indoor marble. The results indicate that the system operation will be hardly affected by illumination, as long as the ambient light can guarantee the accurate focusing of the camera; but it has certain influence on the variation of environment textures, and the response time of initial 3D registration is mainly affected. The experimental results are shown as figure 1. **SBPPS** will not be affected by illumination or texture environment, so the multi-environment tests are not performed. The operation effect is shown as figure 2.

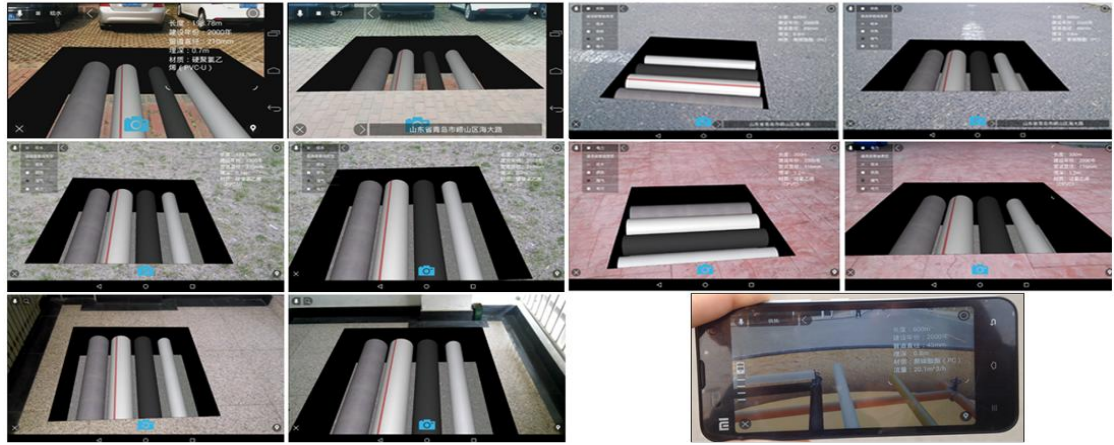

**Fig. 1.** Results of tests on multiple pavements: (1) floor tile covered parking lot; (2) bituminous street; (3) lawn; (4) red floor tile; (5) indoor marble. **Fig. 2.** SBPPS Operation Effect

## 5  Conclusion

Gerhard Schall's work inspired our research [11]. The 7-point likert questionnaire method is adopted, and users include personnel in the public facilities GIS field and outdoor working staff. A series of problems about application in mobile terminal are put forward and need to be solved, including 3d modeling, registration and tracking, application interaction, etc.[53-58].

Based on ARGIS framework, this paper utilizes the Qingdao World Horticultural Expo underground pipeline data to design and achieve the two systems: MVBPPS and SBPPS; MVBPPS adopts the SFM-based registration and tracking, and uses the method TCNN to optimize the process of 3D feature matching, while SBPPS adopts the multi-sensor processing based registration and tracking; expanded application is achieved based on Google Glass and Google Cardboard. It can be seen from research and analysis in this paper that machine vision and sensor show different advantages in the application research of outdoor augmented reality; with the improvement of intelligent mobile terminal processing capacity and sensor precision, it can be imaged that the application of mobile phone based outdoor augmented geologic information system will exist everywhere, changing people's working mode and lifestyle.

**Acknowledgements.** This research is supported by Innovation Fund for Technology Based Firms, China (No: 14c26211100180) and Qingdao science and technology project of China (No: 14-9-2-12-pt).

## Reference

1. Wang, Ke, et al. "Next generation job management systems for extreme-scale ensemble computing." Proceedings of the 23rd international symposium on High-performance parallel and distributed computing. ACM, 2014.
2. Li, Tonglin, et al. "Distributed Key-Value Store on HPC and Cloud Systems." 2nd Greater Chicago Area System Research Workshop (GCASR). 2013.
3. Tianyun Su, Wen Wang, Zhihan Lv, Wei Wu, Xinfang Li. Rapid Delaunay Triangulation for Random Distributed Point Cloud Data Using Adaptive Hilbert Curve. Computers & Graphics. 2015.
4. Lv Z, Rehman S U, Chen G. WebVRGIS: A P2P Network Engine for VR Data and GIS Analysis [J]. Lecture Notes in Computer Science, 2013.
5. RT A. A Survey of Augmented Reality [J]. Presence Teleoperators & Virtual Environments, 1997, 6(4):355-385.
6. Zlatano va D S. Augmented Reality Technology[R].GIS Report No. 17, Delft, 2002

7. King G R, Piekarski W, Thomas B H. ARVino – Outdoor Augmented Reality visualisation of viticulture GIS data[C]. //Forth IEEE & Acm International Conference on Mixed & Augmented Reality. 2005:52--55.
8. Min S, Mei L, Feizhou Z, et al. Hybrid Tracking for Augmented Reality GIS Registration[C]. //Workshop on Frontier of Computer Science & Technology. IEEE, 2007:139 - 145.
9. Guo Y, Qingyun Y L, Zhang W, et al. Application of augmented reality GIS in architecture [J]. The International Archives of Photogrammetry, Remote Sensing and Spatial Information Sciences, 2008, 37: 331-336.
10. Gethin D, Roberts W, Evans A, et al. The Use of Augmented Reality, GPS and INS for Subsurface Data [J]. Proceedings of the Fig XIII International Congress, 2002.
11. Schall G, Mendez E, Kruijff E, et al. Handheld Augmented Reality for Underground Infrastructure Visualization[C]. //Personal & Ubiquitous Computing. 2008:281-291.
12. Schall G, Mulloni A, Reitmayr G. North-centred orientation tracking on mobile phones[C]. //IEEE International Symposium on Mixed & Augmented Reality. IEEE, 2010:267 - 268.
13. Gerhard Schall, Stefanie Zollmann, Reitmayr G. Smart Vidente: advances in mobile augmented reality for interactive visualization of underground infrastructure [J]. Personal & Ubiquitous Computing, 2012, 17(7):1533-1549.
14. T L, C D, A M, et al. Robust detection and tracking of annotations for outdoor augmented reality browsing.[J]. Computers & Graphics, 2011, 35(4):831–840.
15. Schall G, Mulloni A, Reitmayr G. North-centred orientation tracking on mobile phones[C]. //IEEE International Symposium on Mixed & Augmented Reality. IEEE, 2010:267 - 268.
16. Behzadan A H, Kamat V R. Interactive Augmented Reality Visualization for Improved Damage Prevention and Maintenance of Underground Infrastructure [J]. American Society of Civil Engineers, 2014, (339).
17. Xiaoming Li, Zhihan Lv, Jinxing Hu, Ling Yin, Baoyun Zhang, Shengzhong Feng. Virtual Reality GIS Based Traffic Analysis and Visualization System. Advances in Engineering Software. 2015.
18. Zhang, Xu, et al.. Spike-based indirect training of a spiking neural network-controlled virtual insect. 2013 IEEE 52nd Annual Conference on Decision and Control (CDC). IEEE, 2013.
19. Wang, Ke, et al. "Towards Scalable Distributed Workload Manager with Monitoring-Based Weakly Consistent Resource Stealing." (2015).
20. Zhihan Lv, Tengfei Yin, Yong Han, Yong Chen, and Ge Chen. WebVR——web virtual reality engine based on P2P network. Journal of Networks. 6, no. 7 (2011): 990-998.
21. Jiachen Yang, Bobo Chen, Jianxiong Zhou, Zhihan Lv. A portable biomedical device for respiratory monitoring with a stable power source. Sensors. 2015.
22. Shuping Dang, Jiahong Ju, Matthews, D., Xue Feng, Chao Zuo. Efficient solar power heating system based on lenticular condensation. Information Science, Electronics and Electrical Engineering (ISEEE), 2014 International Conference on. April 2014.
23. Wang, Ke, et al. "Overcoming Hadoop Scaling Limitations through Distributed Task Execution."
24. Zhang, Su, Xinwen Zhang, and Xinming Ou. "After we knew it: empirical study and modeling of cost-effectiveness of exploiting prevalent known vulnerabilities across iaas cloud." Proceedings of the 9th ACM symposium on Information, computer and communications security. ACM, 2014.
25. Wei Gu, Zhihan Lv, Ming Hao. Change detection method for remote sensing images based on an improved Markov random field. Multimedia Tools and Applications. 2016.
26. Zhihan Lu, Chantal Esteve, Javier Chirivella and Pablo Gagliardo. A Game Based Assistive Tool for Rehabilitation of Dysphonic Patients. 3rd International Workshop on Virtual and Augmented Assistive Technology (VAAT) at IEEE Virtual Reality 2015 (VR2015), Arles, France, IEEE, 2015.
27. Zhanwei Chen, Wei Huang, Zhihan Lv. Uncorrelated Discriminant Sparse Preserving Projection Based Face Recognition Method. Multimedia Tools and Applications. 2016.
28. Yancong Lin, Jiachen Yang, Zhihan Lv, Wei Wei, Houbing Song. A Self-Assessment Stereo Capture Model Applicable to the Internet of Things. Sensors. 2015.
29. Zhihan Lv, Alaa Halawani, Shengzhong Feng, Haibo Li, and Shafiq Ur Réhman. 2014. Multimodal Hand and Foot Gesture Interaction for Handheld Devices. ACM Transactions on Multimedia Computing, Communications, and Applications (TOMM). 11, 1s, Article 10 (October 2014), 19 pages.
30. Wei Ou, Zhihan Lv, Zanfu Xie. Spatially Regularized Latent topic Model for Simultaneous object discovery and segmentation. The 2015 IEEE International Conference on Systems, Man, and Cybernetics (SMC2015).
31. Wang, Ke, et al. "Using Simulation to Explore Distributed Key-Value Stores for Exascale System Services." 2nd Greater Chicago Area System Research Workshop (GCASR). 2013.

32. Yi Wang, Yu Su, Gagan Agrawal. A Novel Approach for Approximate Aggregations Over Arrays. In Proceedings of the 27th international conference on scientific and statistical database management, ACM, 2015.
33. Zhihan Lv, Alaa Halawani, Shengzhong Feng, Shafiq ur Rehman, Haibo Li. Touch-less Interactive Augmented Reality Game on Vision Based Wearable Device. Personal and Ubiquitous Computing. 2015.
34. Jiachen Yang, Shudong He, Yancong Lin, Zhihan Lv. Multimedia cloud transmission and storage system based on internet of things. Multimedia Tools and Applications. 2016.
35. Yu Su, et al.. In-situ bitmaps generation and efficient data analysis based on bitmaps. In Proceedings of the 24th International Symposium on High-Performance Parallel and Distributed Computing, pp. 61-72. ACM, 2015.
36. Heidemann G, Bax I, Bekel H. Multimodal interaction in an augmented reality scenario[C]//Proceedings of the 6th international conference on Multimodal interfaces. ACM, 2004: 53-60.
37. Ismail A W, Sunar M S. Multimodal Fusion: Gesture and Speech Input in Augmented Reality Environment [J]. Advances in Intelligent Systems & Computing, 2015.
38. Zhou, Xiaobing, et al. "Exploring Distributed Resource Allocation Techniques in the SLURM Job Management System." Illinois Institute of Technology, Department of Computer Science, Technical Report (2013).
39. Li, Tonglin, et al. "ZHT: A light-weight reliable persistent dynamic scalable zero-hop distributed hash table." Parallel & Distributed Processing (IPDPS), 2013 IEEE 27th International Symposium on. IEEE, 2013.
40. Georg Klein and David Murray. Parallel tracking and mapping on a camera phone. In Proceedings of IEEE and ACM International Symposium on Mixed and Augmented Reality, 2009, 83~86.
41. D.P. Robertson and R. Cipolla. Structure from Motion. Practical Image Processing and Computer Vision, John Wiley, 2009.
42. David G. Lowe. Distinctive image features from scale-invariant keypoints. Int. J. Comput. Vision, 60(2):91–110, 2004.
43. M. Muja and D.G Lowe. Fast approximate nearest neighbors with automatic algorithm configuration. In Proc. Int. Conf. . Computer Vision Theory and Application (VISSAPP'09), pages 331–340, 2009.
44. R. Hartley and A. Zisserman. Multiple View Geometry in Computer Vision. Cambridge University Press, 2003.
45. M. Brown and D.G. Lowe. Automatic panoramic image stitching using invariant features. Int. J. Comput. Vision, 2007.
46. S. Yousefi, F. Kondori, and H. Li. 3d gestural interaction for stereoscopic visualization on mobile devices. In 14 Computer Analysis of Images and Patterns (CAIP), pages 555–562, 2011.
47. Milette G, Stroud A. Professional Android Sensor Programming [J]. Wrox, 2012.
48. Ruiz-Ruiz A J, Lopez-de-Teruel P E, Canovas O. A multisensor LBS using SIFT-based 3D models[C]. //International Conference on Indoor Positioning & Indoor Navigation. IEEE, 2012:1 - 10.
49. Castle R, Klein G, Murray D W. Video-rate localization in multiple maps for wearable augmented reality[C]. //In Proc IEEE Int Symp on Wearable Computing, Pittsburgh Pa. 2008:15--22.
50. Chen L, Aihara K. Chaotic simulated annealing by a neural network model with transient chaos[J]. Neural Networks, 1994.
51. JJ H, DW T. "Neural" computation of decisions in optimization problems[J]. Biological Cybernetics, 1985, 52(3):141-152.
52. Li X, Chen D. Augmented reality in e-commerce with markerless tracking[C]. //IEEE International Conference on Information Management & Engineering. IEEE, 2010:609 - 613.
53. Wang, Ke, et al. "Optimizing load balancing and data-locality with data-aware scheduling." Big Data (Big Data), 2014 IEEE International Conference on. IEEE, 2014.
54. Zhao, Dongfang, et al. "FusionFS: Toward supporting data-intensive scientific applications on extreme-scale high-performance computing systems." Big Data (Big Data), 2014 IEEE International Conference on. IEEE, 2014.
55. Yang, Ying, et al. "A steganalytic algorithm for 3D polygonal meshes." Image Processing (ICIP), 2014 IEEE International Conference on. IEEE, 2014.
56. Yang, Ying, Norbert Peyerimhoff, and Ioannis Ivrissimtzis. "Linear correlations between spatial and normal noise in triangle meshes." Visualization and Computer Graphics, IEEE Transactions on 19.1 (2013): 45-55.
57. Yang, Ying, and Ioannis Ivrissimtzis. "Polygonal mesh watermarking using Laplacian coordinates." Computer Graphics Forum. Vol. 29. No. 5. Blackwell Publishing Ltd, 2010.
58. Dingde Jiang, Xu Ying, Yang Han, Zhihan Lv. Collaborative Multi-hop Routing in Cognitive Wireless Networks. Wireless Personal Communications. 2015.